\documentclass[letterpaper]{article} 
\usepackage{aaai23}  
\usepackage{times}  
\usepackage{helvet}  
\usepackage{courier}  
\usepackage[hyphens]{url}  
\usepackage{graphicx} 
\usepackage{natbib}  
\usepackage{caption} 
\usepackage{algorithm}
\usepackage{algorithmic}
\usepackage{amsmath}
\usepackage{amssymb}
\usepackage{subcaption}
\usepackage{multirow}
\usepackage{booktabs}

\usepackage{newfloat}
\usepackage{listings}
\DeclareCaptionStyle{ruled}{labelfont=normalfont,labelsep=colon,strut=off} 
\floatstyle{ruled}
\newfloat{listing}{tb}{lst}{}
\floatname{listing}{Listing}
\DeclareMathOperator*{\argmin}{arg\,min}
\usepackage{hyperref}

\title{Multi-Outputs Is All You Need For Deblur}
\author{
    Sidun Liu,
    Peng Qiao\footnote{Peng Qiao is the corresponding author},
    Yong Dou
}
\affiliations{
    Science and Technology on Parallel and Distributed Laboratory\\
    National University of Defense Technology, Hunan, China\\
    \{liusidun,pengqiao,yongdou\}@nudt.edu.cn

}

\usepackage{bibentry}

\begin{document}

\maketitle

\begin{abstract}
Image deblurring task is an ill-posed one, where exists infinite feasible solutions for blurry image. Modern deep learning approaches usually discard the learning of blur kernels and directly employ end-to-end supervised learning. Popular deblurring datasets define the label as one of the feasible solutions. However, we argue that it's not reasonable to specify a label directly, especially when the label is sampled from a random distribution. Therefore, we propose to make the network learn the distribution of feasible solutions, and design based on this consideration a novel multi-head output architecture and corresponding loss function for distribution learning. Our approach enables the model to output multiple feasible solutions to approximate the target distribution. We further propose a novel parameter multiplexing method that reduces the number of parameters and computational effort while improving performance. We evaluated our approach on multiple image-deblur models, including the current state-of-the-art NAFNet. The improvement of best overall (pick the highest score among multiple heads for each validation image) PSNR outperforms the compared baselines up to \textbf{0.11$\sim$0.18dB}. The improvement of the best single head (pick the best-performed head among multiple heads on validation set) PSNR outperforms the compared baselines up to \textbf{0.04$\sim$0.08dB}. 
The codes are available at \href{https://github.com/Liu-SD/multi-output-deblur}{https://github.com/Liu-SD/multi-output-deblur}.
\end{abstract}

\section{1\quad Introduction}

Image restoration, aiming at reconstructing a high-quality image from its degraded counterpart, is one of the fundamental problems in computer vision. Image deblurring is one of these tasks, where degradations are caused by the motion or jittering of camera or objects. The task of image deblurring is to remove the blurry trace and restore the sharp image. When the exposure time is not short enough, the image records the instantaneous movement of the objects, resulting in the blurry one. 

With the help of the powerful fitting ability of CNN-based and recently popular self-attention-based~\cite{vaswani2017attention,dosovitskiy2020image} models, the supervised image deblurring achieves tremendous improvement~\cite{dong2014learning,qiao2017learning,chen2021pre,liang2021swinir,chen2021hinet,park2020multi,zamir2022restormer,wang2022uformer,chu2021tlc}. 
MPRNet~\cite{zamir2021multi} uses a multi-stage training strategy to progressively restore a degraded image; HiNet~\cite{chen2021hinet} uses a Half instance normalization layer to boost the restoration; TLC~\cite{chu2021tlc} restricts the statistics aggregation to local at test phase; Restormer~\cite{zamir2022restormer} and MAXIM~\cite{tu2022maxim} put the self-attention to feature dimension to relief the computing complexity; NAFNet~\cite{chen2022simple} combines the MobileNet-style~\cite{howard2017mobilenets} convolution module and channel attention~\cite{hu2018squeeze} to make the model lightweight but powerful.
The sophisticated architecture of these networks gives the models strong representation ability, allowing them to excel at the deblurring task. In supervised learning, we consider a blurry image as a certain stacking of a frame sequence, the corresponding label image is usually defined as the middle frame in this sequence. However, due to the random jittering of camera or objects during the image capturing, the middle frame is still with uncertainty. There may exist several pixel shifts between the label image and the restored image. When training models with respect to MSE-like loss function, the restored images lose the sharp edges and meaningful textures.

\begin{figure*}[t]
    \centering
    \includegraphics[width=0.9\textwidth]{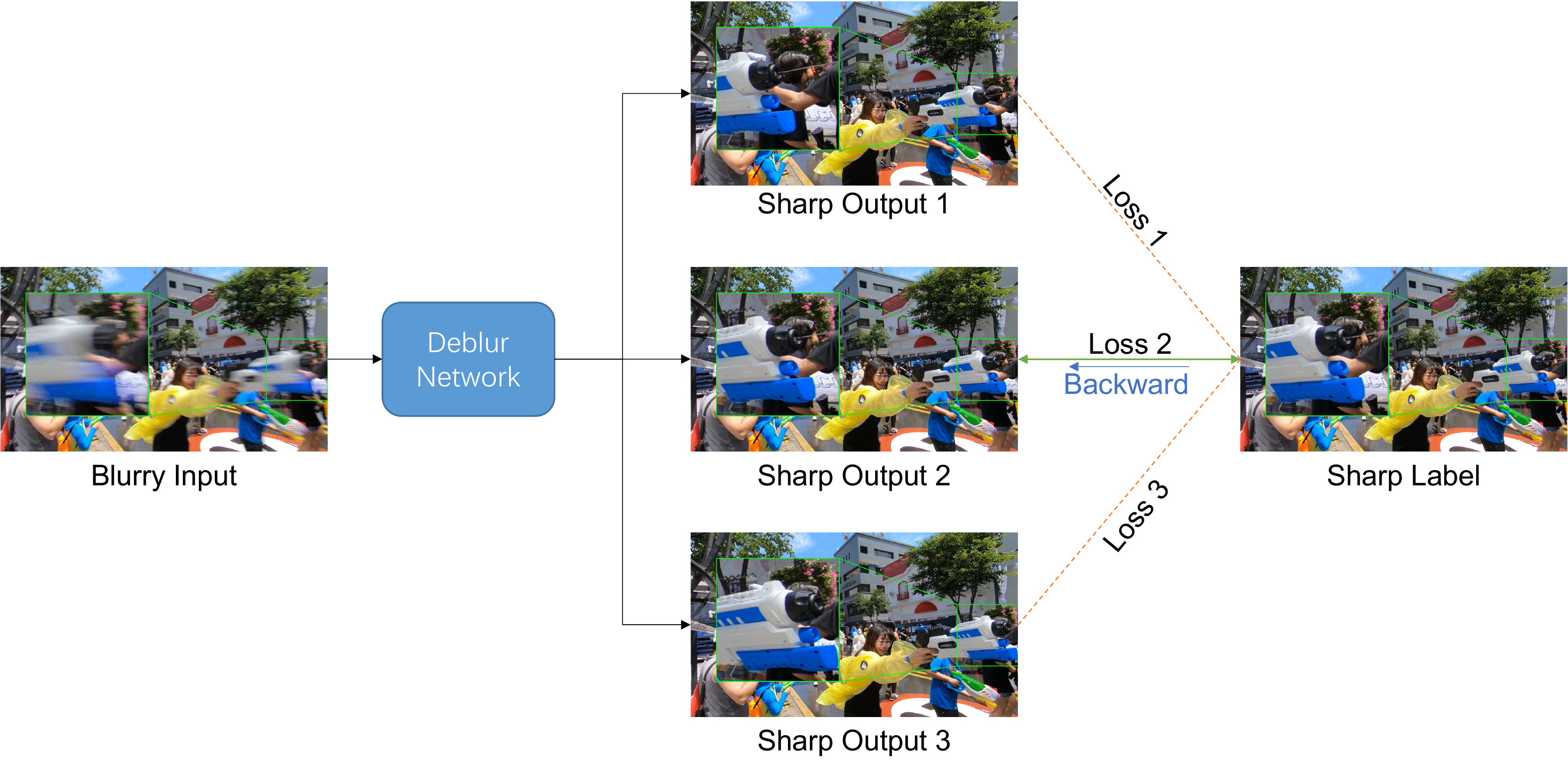}
    \caption{The training flow of the proposed multi-outputs approach. The deep neural network generates multiple sharp outputs. But only the output with the lowest loss is used for back-propagation.}
    \label{fig:main}
\end{figure*}


In a supervised deblurring task, we assume that the corresponding label images are samples from a distribution but not ones defined above. The samples in this distribution are all feasible restoration results. Training the models using the MSE-like loss function can only capture the expectation of this distribution, which limits the quality of the restored images. 

In this scenario, it is better to train the model to learn the distribution, instead of learning its expectation. However, learning the distribution directly is difficult \cite{kingma2013vae}. 
We propose to divide the distribution into several clusters, supervising the model to learn the expectations of each cluster. 
Inspired by k-means~\cite{lloyd1982least}, we incorporate an EM algorithm to optimize the model to capture the distribution. At the beginning, the cluster centers are randomly initialized. For E-step, each label is assigned to the nearest cluster, and for M-step, the cluster centers are updated to be the expectations of their corresponding clusters. For implementation, we design a multi-head output layer to generate clustering centers. As only the output layer and supervision manner are modified, our proposed method can be easily extended to existing models.

The more clusters, the better the distribution is fitted. But directly increasing the number of output heads is not acceptable. On the one hand, the parameters and computations increase linearly as the number of clusters increasing. On the other hand, some heads may not be sufficiently utilized. We observe that there are correlations between the output images from different heads. As described above, the transformation from sharp images to a blurry image is additive. Therefore, we combine the heads of model output in pairs to obtain the extended multi-head output. In this way, the parameters and computations of the last layer are the square roots of the number of clusters. Meanwhile, the shared parameters will enhance the connections between different heads.



The experiments are performed on multiple models, including the NAFNet~\cite{chen2022simple}, Restormer~\cite{zamir2022restormer}, HiNet~\cite{chen2021hinet}, and MPRNet~\cite{zamir2021multi}, trained on GoPro~\cite{Nah_2017_CVPR} and validated on various datasets. The experiments on NAFNet-width32~\cite{chen2022simple} show that with our multi-head extension, the best overall (pick the highest score among multiple heads for each validation image) PSNR improves 0.05$\sim$0.22dB with a different number of heads, and the best single head (pick the best-performed head among multiple heads on validation set) PSNR outperforms it up to 0.2dB. When the head combination is enabled, the experimented models achieve 0.15dB higher best overall PSNR and 0.05dB higher best single head PSNR on average, compared to single head baselines of the models above. The multi-head NAFNet-width64 with head combination achieves 33.82~(+0.11dB) best overall PSNR, and 33.75~(+0.04dB) best single head PSNR, which exceeds the state-of-the-art.

We further analyze the multi-head outputs and find that with our semi-supervised training strategy, multiple feasible but different results are generated. The diversity of the multi-headed outputs are reflected in the blurry regions of the image.

Our contributions can be summarized as follows:

\begin{itemize}
    \item We analyze the reason that limits image recovery performance in supervised image deblurring tasks. Since deblurring is ill-posed, using labeled image supervision will likely result in the model outputting images out of the distribution.
    \item We point out that the distribution of sharp images should be learned, not their expectations. And a semi-supervised learning strategy based on the EM algorithm is proposed to learn multiple clusters to approximate the distribution.
    \item We further propose to combine the output heads in pairs such that the parameters and computations of the output layer is the square root of the number of clusters. The connection between the clusters is also enhanced by sharing parameters.
    \item Our approach can be simply extended to existing models. We use NAFNet and other mainstream models as backbones for our experiments. The experiments demonstrate the effectiveness of the proposed method. Not only does the best overall PSNR exceed the baselines, but also the PSNR of each head is comparable to or better than them.
\end{itemize}

\section{2\quad Related Work}

\subsection{2.1\quad Image Deblurring}

The deep learning methods have achieved significant success in image deblurring and other low-level vision tasks such as image denoise~\cite{tian2020deep}, image deraining~\cite{li2019single}, and image super-resolution~\cite{yang2019deep}. Early works propose to estimate the blur kernel~\cite{chakrabarti2016neural,ren2020neural,schuler2015learning,sun2015learning,tran2021explore}. But, since the characteristics of blur are complex, the blur kernel estimation method is not practical in real scenarios. DeepDeblur~\cite{Nah_2017_CVPR} gives up estimating blur kernel but directly maps a blurry image to its sharp counterpart. Following this paradigm, a series of methods~\cite{zamir2021multi,chen2021hinet,zamir2022restormer,chu2021tlc} constantly refresh the SOTA. At present, the NAFNet~\cite{chen2022simple} achieves highest PSNR~(33.71 dB) on the GoPro~\cite{Nah_2017_CVPR} image deblur dataset.

\subsection{2.2\quad Deblurring Datasets}

The commonly used benchmark for image deblurring is GoPro~\cite{Nah_2017_CVPR}, where the input blurry images are synthetic. The GoPro dataset takes 240 fps videos with a GOPRO camera and then averages varying numbers (7 - 13) of successive latent frames to produce blurs of different strengths. After averaging, the integrated signal is then transformed into pixel value by nonlinear CRF (Camera Response Function). 

HIDE~\cite{shen2019human} is a dataset for human-ware image deblurring, where the images consist of densely annotated foreground human bounding boxes. The blurry images from the HIDE are synthetic in the same way as GoPro.


The ill-posed nature of deblurring makes every sharp image corresponding to each blurry one a feasible solution. For the convenience of supervision, both datasets define the middle frame among the sharp frames as the corresponding sharp image. However, due to the random jittering of camera or objects during the image capturing, the middle frame is stochastic, forming a distribution. Such randomness makes the model impossible to predict the exact middle frame for the input blurry image.

\subsection{2.3\quad Distribution Learning}

To learn the distribution, we have to assume it belongs to a restricted family of distributions. Gaussian mixture model~(GMM)~\cite{Reynolds2009} is a distribution estimation model with an assumption that the data are sampled from multiple Gaussian distributions. An EM algorithm exists to estimate the expectation and covariance matrix for the GMM. A special case is derived when the hard partition is made. Then the GMM degrades to the k-means~\cite{lloyd1982least} problem. 

In the deblurring task, the restoration result follows an unknown distribution $P(y|x)$. When the distribution is approximated by the mixtures of Gaussian distribution with hard partition, the k-means-like algorithm can be applied to learn the distribution.



\section{3\quad Approach}

In this section, we provide more detailed explanations about the multi-head training strategy and the head combination method in the following subsections. 

\subsection{3.1\quad Multi-Head Training}

Image restoration, e.g. deblurring, is an ill-posed problem as there exists infinite feasible solutions. We expect the learned model to generate one of these feasible solutions. However, when the supervised training is applied, the model is forced to generate the one which exactly matches the target sharp image. If the selection of the target sharp image is deterministic, the model is able to learn to generate the restored image. However, in fact, the selection of target sharp images is stochastic. E.g., the popular single image deblurring dataset GoPro~\cite{Nah_2017_CVPR} synthesizes the input blurry image with a picture sequence and selects the middle frame of that sequence as the label. The random jittering of camera or objects during the capturing still causes the randomness to the target sharp image. If we assume the label is sampled from an unknown distribution, the model trained in a supervised manner is tend to generate the expectation of such distribution. However, the expectation solution may provide inferior visual quality. In fact, for the motion deblurring task, there exist several pixel shifts between feasible solutions. So the expectation of such distribution loses sharp edges and meaningful textures, making it far from feasible solutions.

To tackle this problem, we point out that the distribution of the target sharp image should be learned, instead of its expectation. 
However, without any prior about the distribution, learning the distribution directly is difficult. Therefore, we divide the distribution into several clusters, represented by their cluster centers. When the cluster number is large enough, the distribution can be approximated by these centers. 
Then we supervise the model to generate these cluster centers. 

Given a set of image pairs 
$ \{ (x_i, y_i)\}_{i=1}^n $, the model generates $K$ cluster centres 
$\{ \mu_j(x) \}_{j=1}^K$ for each input image $x$. We aim to assign the label $y$ to one of these clusters, so as to minimize the within-cluster sum of squared errors. Formally, the objective is to find assignment $S$ and cluster centers generator $\mu$ as Equation~\ref{definition}.

\begin{equation}
    \label{definition}
    \argmin_{\mathbf{S}, \mu} \sum_{i=1}^K \sum_{(x,y)\in \mathbf{S}_i} ||y-\mu_i(x)||^2
\end{equation}

This problem is NP-hard. Inspired by k-means, we designed an EM algorithm to heuristically optimize this objective in Equation~\ref{definition}. Given an initial cluster centers generator $\mu^{(1)}$, the algorithm proceeds by alternating between two steps:

\textbf{Assignment step:} Assign each label $y$ to the cluster with the nearest cluster centres (Equation~\ref{assignment}).

\begin{multline}
    \label{assignment}
    \mathbf{S}_i^{(t)} = \{ (x,y): ||y-\mu^{(t)}_i(x)||^2\leq ||y-\mu^{(t)}_j(x)||^2   \\
     \forall j, 1\leq j\leq K \}
\end{multline}

\textbf{Update step:} Update each cluster centre generator $\mu^{(t)}$ with its assigned image pairs in ${S}^{(t)}$.

\begin{equation}
    \label{update}
    \mu_i^{(t+1)} = \min \sum_{(x,y)\in \mathbf{S}_i^{(t)}} ||y-\mu_i^{(t)}(x)||^2
\end{equation}

In practice, the deep neural network (DNN) with multi-head outputs is used as a generator $\mu$. Each head $\mu_i$ is parameterized by $\theta_i$. The assignment and update step with DNN is transformed into:

\textbf{Assignment step with DNN:} Find the head $\mu_{\theta_{\hat{k}}}$ with the minimum loss $L$ with respect to the label $y$. 

\begin{equation}
    \label{assignment_nn}
    \hat{k} = \argmin_i L(\mu_{\theta_i}(x), y)\quad 1\leq i\leq K
\end{equation}
$L$ is the MSE-like loss, e.g. Charbonnier loss~\cite{charbonnier1994two} or PSNR loss.

\textbf{Update step with DNN:} Update the parameter $\theta_{\hat{k}}$ of head $\mu_{\theta_{\hat{k}}}$ to get one-step closer to $y$.

\begin{equation}
    \label{update_nn}
    \mu_{\theta_{\hat{k}}} = \mu_{\theta_{\hat{k}}} - \alpha\nabla L(\mu_{\theta_{\hat{k}}}(x), y)
\end{equation}
$\alpha$ is the learning rate.




\begin{figure}
    \centering
    \includegraphics[width=0.7\columnwidth,trim={0.5cm 0.5cm 0.5cm 0.5cm},clip]{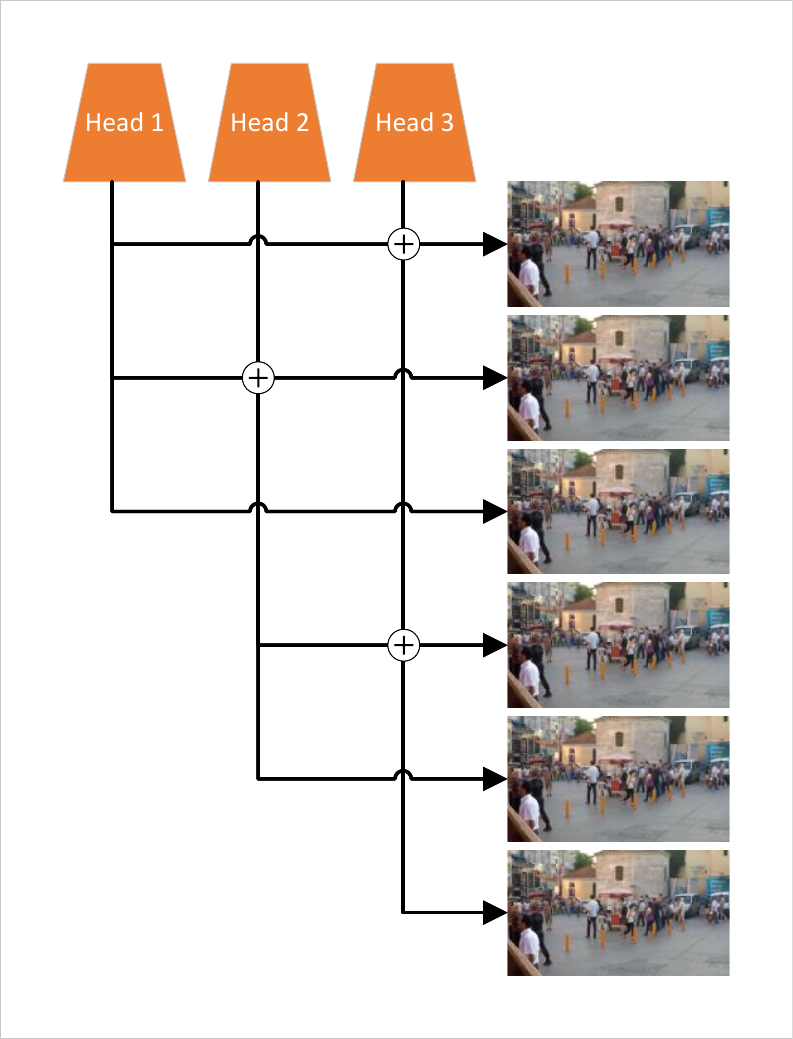}
    \caption{An example of head combination method when the number of heads is 3. The multi-head outputs are combined in pairs (including themselves) to obtain the extended outputs.}
    \label{fig:multi-head}
\end{figure}

\subsection{3.2 \quad Head Combination}
\begin{figure*}
    \centering
    \includegraphics[width=1\textwidth]{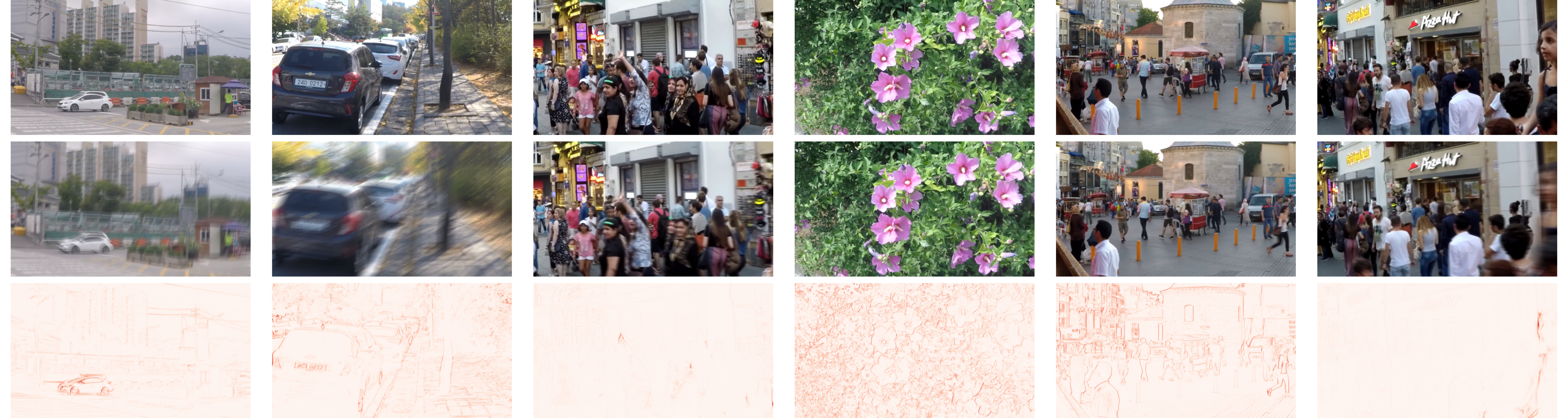}
    \caption{A heatmap visualization to show the residuals between different heads. A trained 4-head NAFNet-width32 with the head combination is used for illustration. The first row is the sharp image, and the second row is the corresponding blurry image. The heatmaps in the third row are computed by accumulating and normalizing the absolute residuals for all pairs of extended heads. Differences between outputs are mainly in the blurry regions, especially at their edges.}
    \label{fig:heatmaps}
\end{figure*}
\begin{figure}[t]
    \centering
    \includegraphics[width=0.75\columnwidth]{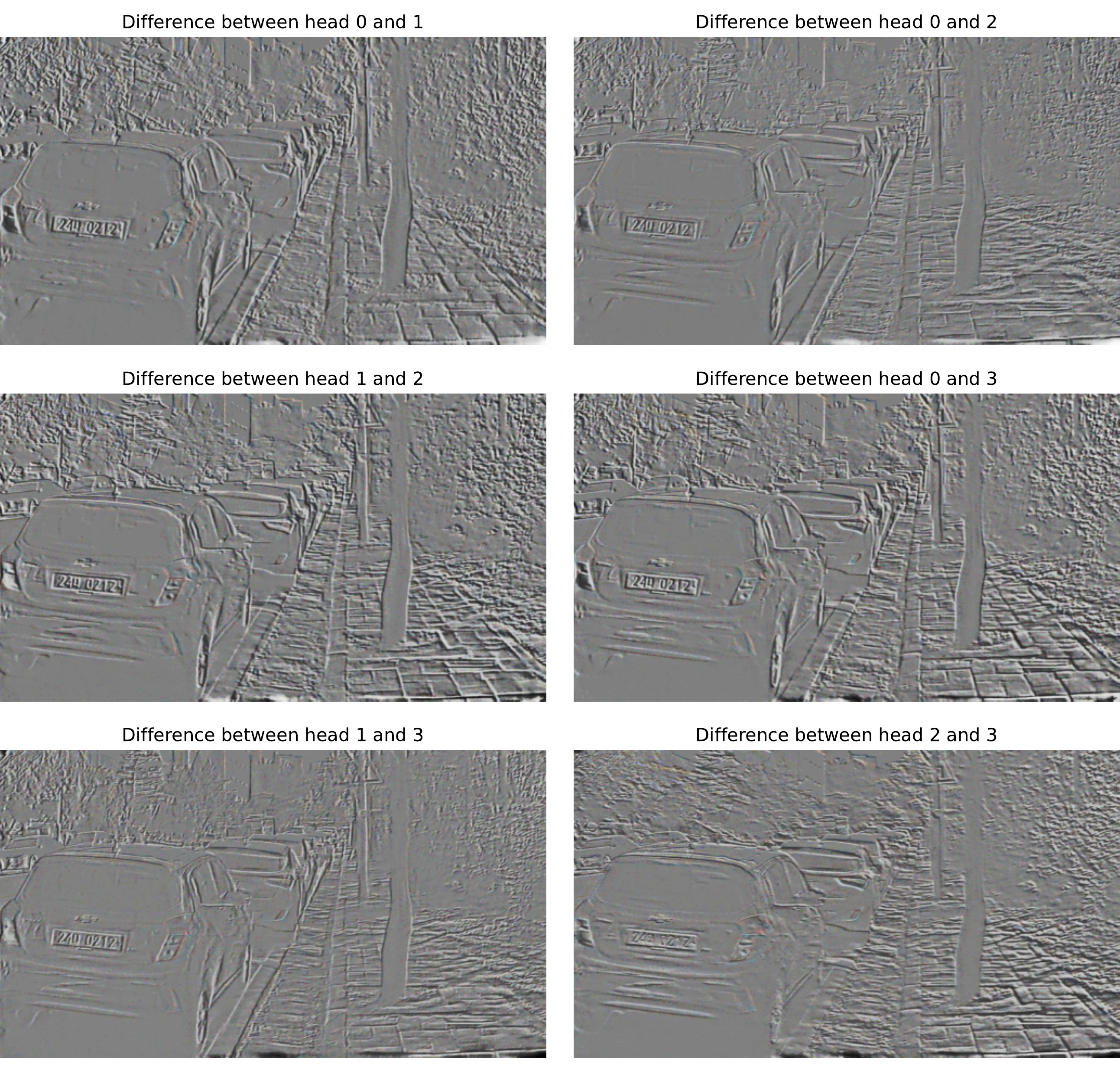}
    \caption{An visualization to show the pixel shifts between different heads. A trained 4-head NAFNet-width64 is used for illustration. For the difference between head $i$ and $j$, if the pixel value of $i$ is larger than that of $j$, the corresponding position becomes bright, and vice versa. An object's left edge is bright and right edge is dark means that this object shifts left from output $i$ to output $j$.}
    \label{fig:diff_between_heads}
\end{figure}




In the previous section, we described the proposed approach to approximate the distribution of target sharp images with multiple cluster centers. In general, more cluster centers may lead to greater performance gains. However, in practice, this causes some problems. First, the parameters and computations of the output layer grow linearly to the number of heads. Second, as we only update the nearest head to label, the heads with relatively worse performance have less chance to be updated, which means some heads can't be sufficiently utilized. 

We notice as in Figure~\ref{fig:heatmaps} that there are correlations between the multi-head outputs. Some slight shifts occur between heads. We assume the input blurry image is a certain stack of a frame sequence, then the processing done by each head is to erase with a residual network architecture all frames but the frame to be restored. So the functions of these heads overlap to some extent. These overlaps can be encoded by sets of shared parameters, and the parameters and computations can be reduced.

Based on this assumption, we proposed a head combination method. We traverse all the output heads and combine them in pairs (including themselves) to get the extended multi-head outputs. Then we apply the multi-head loss to them. As the transformation from sharp to a blurry image is additive, we use simple addition as the combination strategy. The head combination is described as Equation~\ref{head_combination}.

\begin{equation}
    \label{head_combination}
    \mu'_{i,j}(x)=\frac{\mu_i(x)+\mu_j(x)}{2}\quad 1\leq i\leq j \leq K
\end{equation}

If the model has $K$ output heads, then the number of extended heads is $K'=K(K+1)/2$. 

With the head combination, the parameters and computations of the last layer are square-root to the number of extended heads. Besides, the correlations between heads are enhanced by sharing parameters. When one of the extended heads is updated, other correlated heads are updated as well, which avoids the problem of insufficient training for certain heads.

\section{4\quad Experiments}

\subsection{4.1\quad Dataset}

We use GoPro~\cite{Nah_2017_CVPR} datasets for evaluation. GoPro is a commonly used motion deblurring dataset that contains 2,103 image pairs for training and 1,111 pairs for evaluation. 

Furthermore, to demonstrate generalizability~\cite{zamir2021multi, wang2022uformer, chen2022simple}, we take the GoPro trained model and directly apply it on the test images of HIDE~\cite{shen2019human} dataset. The HIDE's test set contains 2,025 images, specifically built for human-aware deblurring. 

\subsection{4.2\quad Implementation Details}

We use multiple commonly compared models to evaluate our method, including:

\textbf{MPRNet}~\cite{zamir2021multi} is a multi-stage architecture, which progressively learns restoration functions for the degraded inputs, thereby breaking down the overall recovery process into several steps. As there are multiple outputs from different steps, we adopt the multi-head output layer to all output layers.

\textbf{HINet}~\cite{chen2021hinet} uses a novel Half Instance Normalization Block (HIN Block) to boost the performance of the image restoration network. 

\textbf{Restormer}~\cite{zamir2022restormer} propose a multi-Dconv head transposed attention (MDTA) module to aggregate local and non-local pixel interactions, which is efficient to process high-resolution images. 

\textbf{NAFNet}~\cite{chen2022simple} use a simple U-shaped architecture to achieve the SOTA. 
There are only MobileNet-style~\cite{howard2017mobilenets} convolutions, SE~\cite{hu2018squeeze} channel attentions, and shortcuts on the NAFNet.

For all the models above, we replace the original output layer with our multi-head output layer by expanding the channels of the output layer by a factor of $K$. The hyper-parameters and settings are kept the same as reported in the papers.

We use the base version of NAFNet with latent width of 32 (NAFNet-width32) for ablation study and analysis. All the models are used for overall evaluation. All the experiments are conducted on a server with 8 Tesla V100 GPUs.

\subsection{4.3\quad Ablation Study}

\begin{figure}[t]
    \centering
    \begin{subfigure}[b]{1\columnwidth}
        \centering
        \includegraphics[width=1\columnwidth]{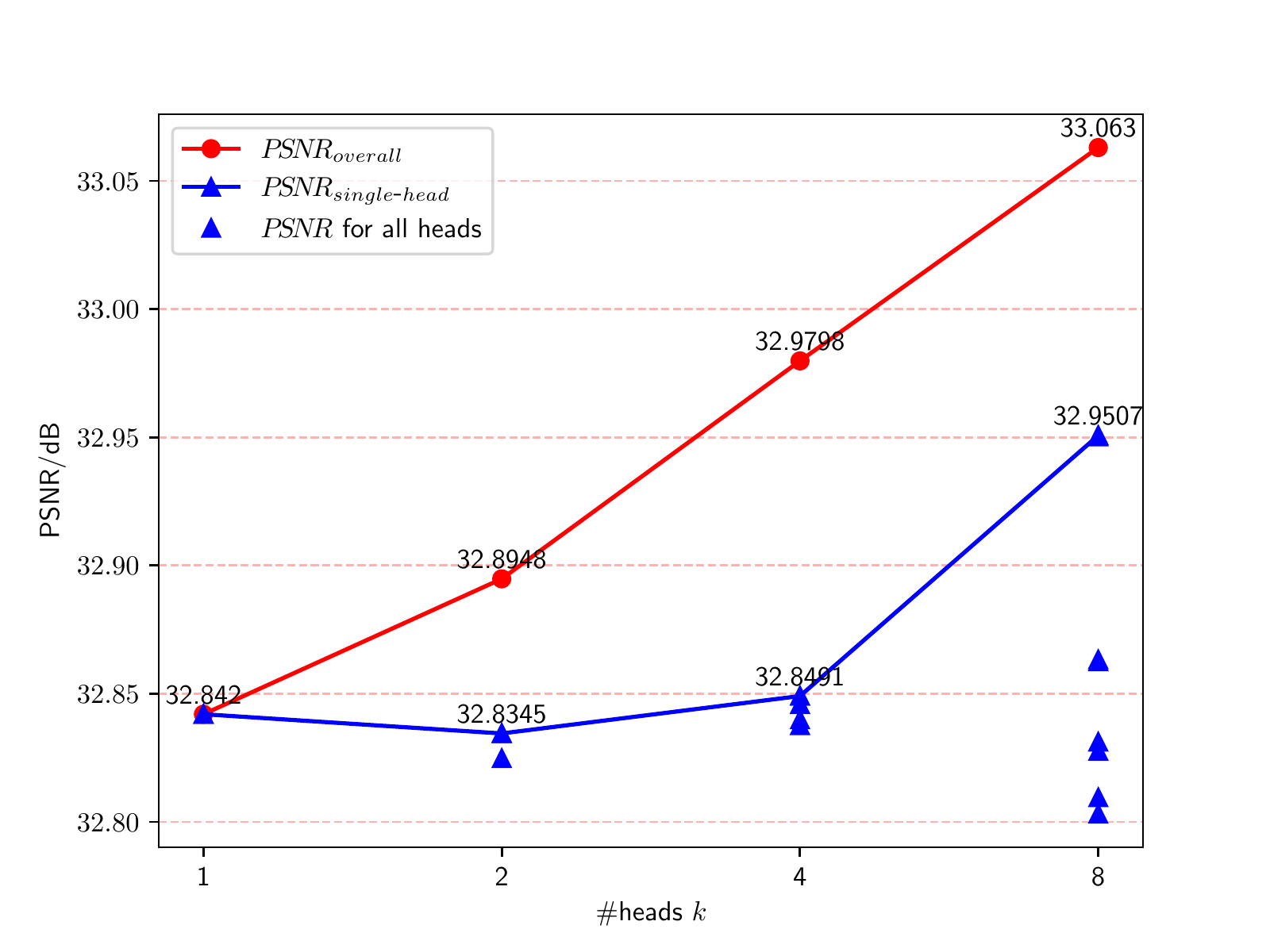}
        \caption{Without the head combination.}
        \label{fig:head_psnr}
    \end{subfigure}
    \begin{subfigure}[b]{1\columnwidth}
        \centering
        \includegraphics[width=1\columnwidth]{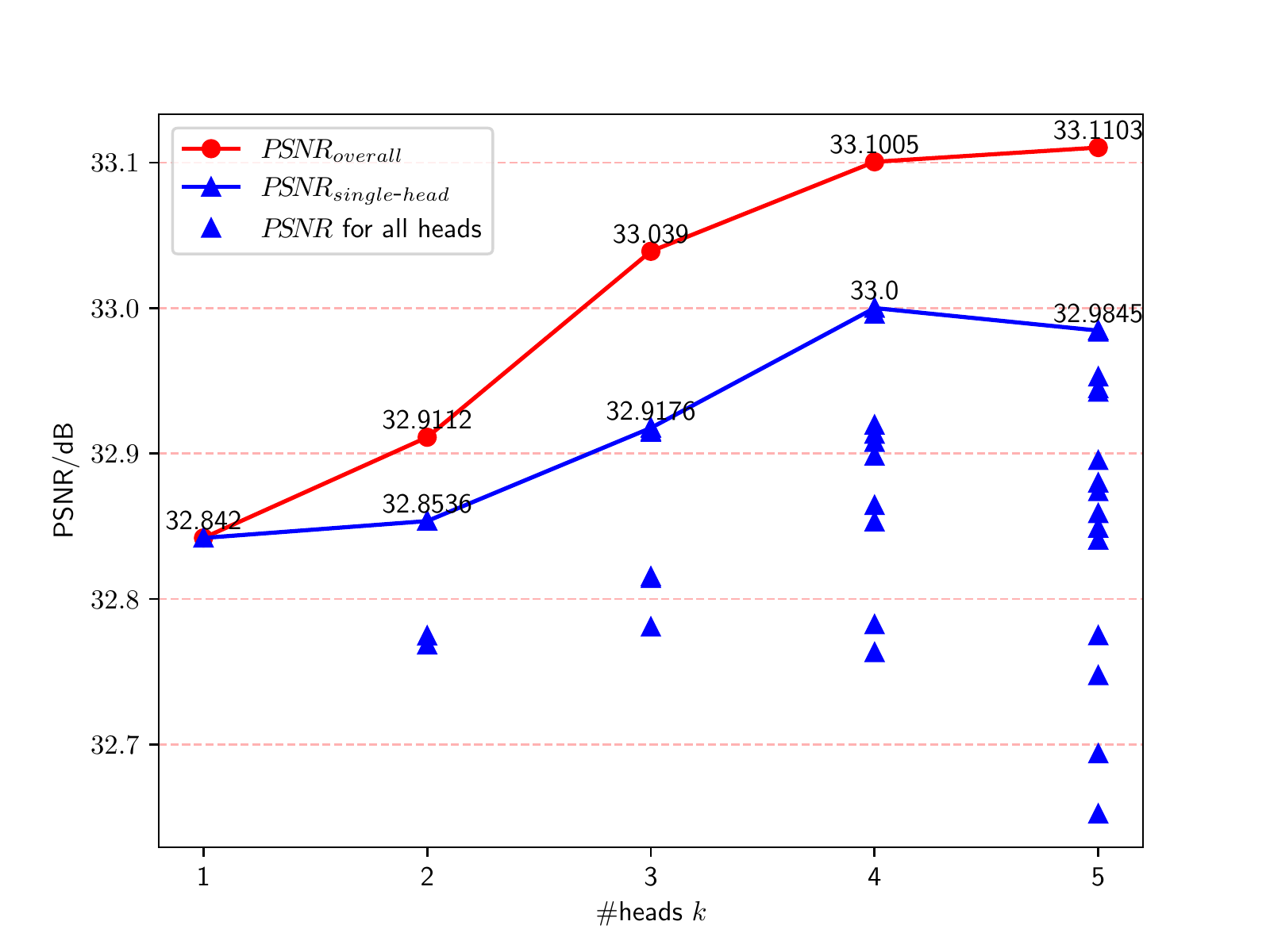}
        \caption{With the head combination.}
        \label{fig:head_psnr_c}
    \end{subfigure}
    \caption{Ablation study on the influence of the number of heads to two types of metrics defined in equation~\ref{metric_overall} and \ref{metric_single_head}. Besides, the PSNRs for all heads are plotted with blue $\blacktriangle$.}
    \label{fig:head_psnrs}
\end{figure}


In this section, we train our multi-head NAFNet-width32 on the GoPro dataset with a variant number of heads. The impact of head combination is also evaluated. There are two types of metrics we used to evaluate our method. The first is the $P\!S\!N\!R_{overall}$, which picks the highest score among multiple heads for each validation image~(Equation \ref{metric_overall}); the second is the $P\!S\!N\!R_{single\mbox{-}head}$, which picks the best performed head among multiple heads on validation set~(Equation \ref{metric_single_head}). Here, the (extended) head number is $K$, and the size of validation set is $N$.



\begin{subequations}
\begin{align}
P\!S\!N\!R_{overall} = \frac{1}{N}\sum_{i=1}^N \max_j P\!S\!N\!R\left(\mu_j(x_i),y_i\right) \label{metric_overall}\\
P\!S\!N\!R_{single\mbox{-}head} = \max_j \frac{1}{N}\sum_{i=1}^N  P\!S\!N\!R\left(\mu_j(x_i),y_i\right) \label{metric_single_head}\\
1\leq j \leq K\notag
\end{align}
\end{subequations}

We show the effects of head numbers on model performance in figure~\ref{fig:head_psnr}. It is observed that not only the $P\!S\!N\!R_{overall}$ but also the $P\!S\!N\!R_{single\mbox{-}head}$ is positively correlated with the number of heads. However, the parameter numbers and computation cost limit the further increment of head numbers. Not all heads perform equally well. A possible explanation is that some heads are in charge of recovering out-of-distribution samples, while other heads are not sufficiently trained. 

Figure~\ref{fig:head_psnr_c} shows the PSNR gains when the head combination is enabled. We can observe that the 4-head model with combination performs better than the 8-head model without combination. On the one hand, a 4-head combination is able to generate 10 outputs. On the other hand, the shared parameters encoded the connections between heads, so that more heads are sufficiently utilized. However, the gains of the $P\!S\!N\!R_{overall}$ slow down and the $P\!S\!N\!R_{single\mbox{-}head}$ drops when the number of heads goes to 5. So we set the head numbers to be 4 and enable the head combination for subsequent experiments.

\subsection{4.4\quad Analysis on Multi-Head Architecture}

\begin{figure}[t]
     \centering
     \begin{subfigure}[b]{0.8\columnwidth}
         \centering
         \includegraphics[width=\textwidth]{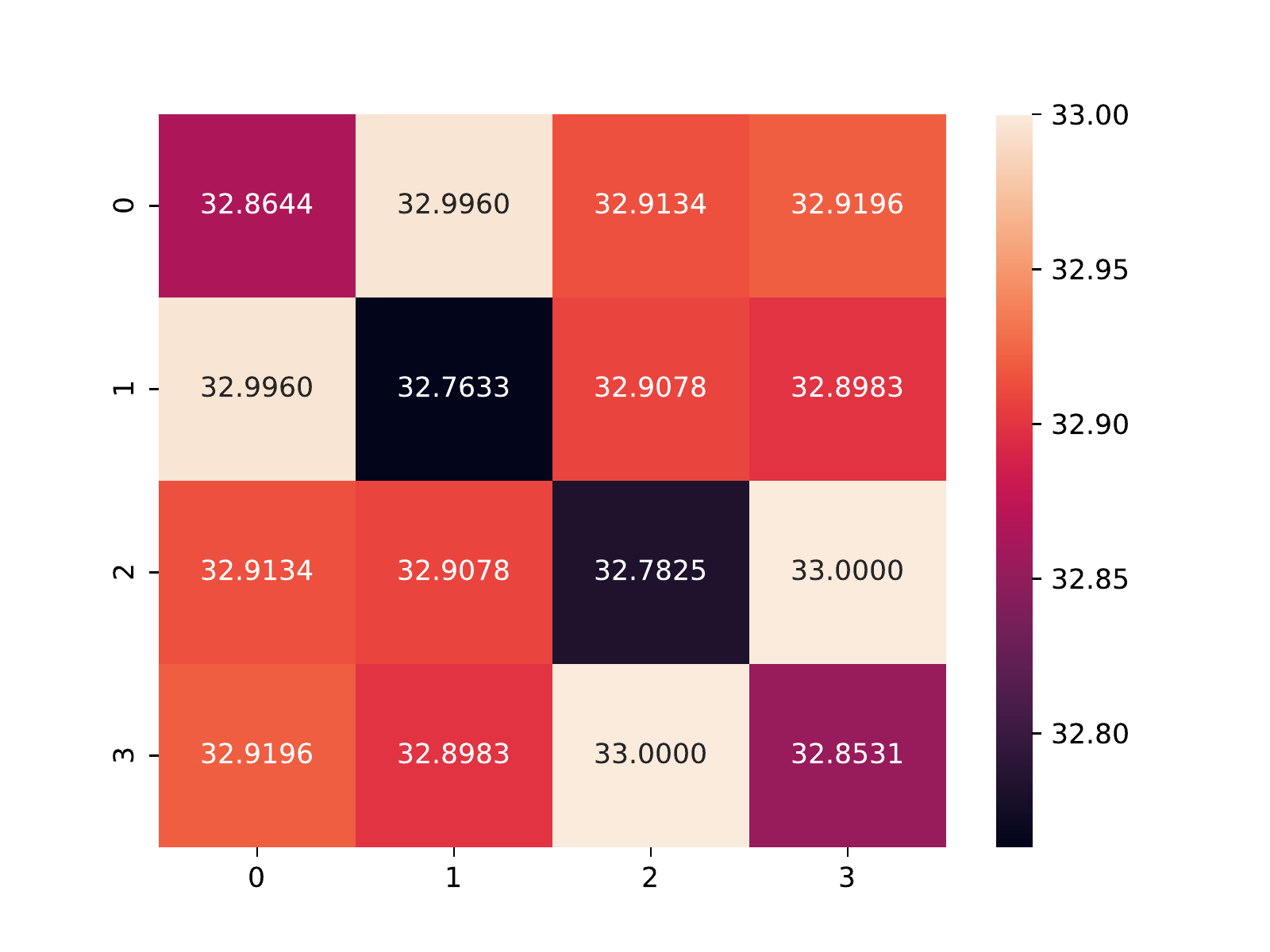}
         \caption{$\#heads=4$}
         \label{fig:psnr_heatmap4}
     \end{subfigure}
     \begin{subfigure}[b]{0.8\columnwidth}
         \centering
         \includegraphics[width=\textwidth]{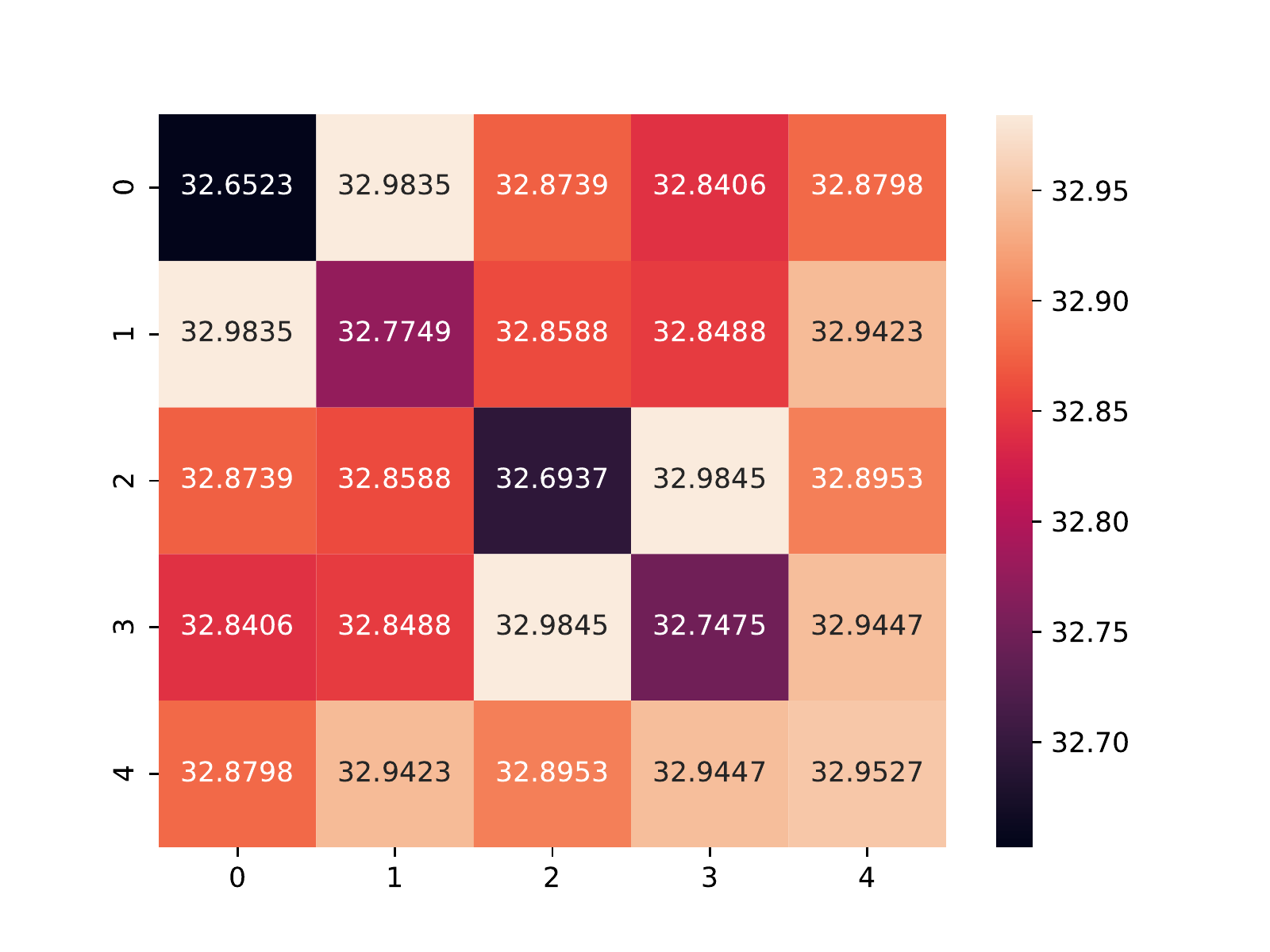}
         \caption{$\#heads=5$}
         \label{fig:psnr_heatmap5}
     \end{subfigure}
        \caption{The PSNR of the output from each pair of heads, using the heatmap format.}
        \label{fig:psnr_heatmap}
\end{figure}

In this section, we analyze the restored sharp images from multiple heads and answer the following questions. 

\textbf{How do multi-head outputs look like?} The differences between the output images are barely observable with human eyes, but the pixel-wise differences are significant. The visualization in figure~\ref{fig:heatmaps} shows that the differences between heads are mainly in the blurry regions, especially the edge of objects. For these blurry regions, there exist infinite solutions, and our trained model generates multiple different solutions. So the heatmap is lighter in the corresponding area. The visualization proves that our trained model is able to generate multiple feasible solutions for the blurry regions, instead of their expectation. One of the outputs matches the label most, but all of them are feasible as a sharp image.

If the camera shakes when taking pictures, the entire image is blurry, so the pixel shift of the entire image should be observed when multi-head output is applied. We use a trained 4-head model without head combination to visualize such a phenomenon in the 2nd picture of figure~\ref{fig:heatmaps}, where the camera shakes violently. The figure shows the normalized residuals between each pair of outputs. A brighter area means the pixel value increase and vice versa. E.g., an object's left edge is bright and the right edge is dark means that this object shifts to the left. We can observe that the edge of all the objects in the picture have similar patterns, proving that there exist pixel shifts between multi-head outputs.

\newcommand\unknowna{\multicolumn{2}{c|}{-}}
\newcommand\unknownb{\multicolumn{2}{c}{-}}
\begin{table*}[t]
\centering
\caption{The results on various motion deblurring datasets. The models with the suffix \textit{-MH-C} are our multi-head variants with the head combination. The models are trained only on the GoPro dataset and directly applied to the HIDE benchmark dataset. $^*$\textit{Single head} for $P\!S\!N\!R_{single\mbox{-}head}/S\!S\!I\!M_{single\mbox{-}head}$ and \textit{Overall} for $P\!S\!N\!R_{overall}/S\!S\!I\!M_{overall}$.}
\label{tab:result}
\begin{tabular}{l|c|ll|ll}
\toprule
\multirow{2}{*}{Method}& \multirow{2}{*}{Metric$^*$} & \multicolumn{2}{c|}{GoPro} & \multicolumn{2}{c}{HIDE} \\ 
&& PSNR & SSIM & PSNR & SSIM \\ \midrule 
Gao \textit{et al.}~\cite{gao2019dynamic}      &           - & 30.90 & 0.935 & 29.11 & 0.913 \\ 
DBGAN~\cite{zhang2020deblurring}               &           - & 31.10 & 0.942 & 28.94 & 0.915 \\ 
MT-RNN~\cite{park2020multi}                    &           - & 31.15 & 0.945 & 29.15 & 0.918 \\ 
DMPHN~\cite{zhang2019deep}                     &           - & 31.20 & 0.940 & 29.09 & 0.924 \\ 
Suin \textit{et al.}~\cite{suin2020spatially}  &           - & 31.85 & 0.948 & 29.98 & 0.930 \\ 
SPAIR~\cite{purohit2021spatially}              &           - & 32.06 & 0.953 & 30.29 & 0.931 \\ 
MIMO-UNet+~\cite{cho2021rethinking}            &           - & 32.45 & 0.957 & 29.99 & 0.930 \\ 
IPT~\cite{chen2021pre}                         &           - & 32.52 &\quad- &\quad- &\quad- \\ 
MAXIM-3S~\cite{tu2022maxim}                    &           - & 32.86 & 0.961 & 32.83 & 0.956 \\ 
\midrule
MPRNet~\cite{zamir2021multi}                   &           - & 32.66 & 0.959 & 30.96 & 0.939 \\ 
\multirow{2}{*}{MPRNet-MH-C~(\textbf{Ours})}   & Single head & 32.65$^{-0.01}$ & 0.959$^{+0.00}$ & 30.95$^{-0.01}$ & 0.940$^{+0.01}$ \\ 
                                               &     Overall & 32.78$^{+0.12}$ & 0.960$^{+0.01}$ & 31.07$^{+0.11}$ & 0.940$^{+0.01}$ \\ 
\midrule
HINet~\cite{chen2021hinet}                     &           - & 32.77 & 0.959 & 30.33 & 0.932 \\ 
\multirow{2}{*}{HINet-MH-C~(\textbf{Ours})}    & Single head & 32.83$^{+0.06}$ & 0.960$^{+0.01}$ & 30.41$^{+0.08}$ & 0.933$^{+0.01}$ \\ 
                                               &     Overall & 32.95$^{+0.18}$ & 0.961$^{+0.02}$ & 30.50$^{+0.17}$ & 0.934$^{+0.02}$ \\ 
\midrule
Restormer~\cite{zamir2022restormer}            &           - & 32.92 & 0.961 & 31.22 & 0.942 \\ 
\multirow{2}{*}{Restormer-MH-C~(\textbf{Ours})}& Single head & 33.01$^{+0.09}$ & 0.962$^{+0.01}$ & 31.36$^{+0.14}$ & 0.944$^{+0.02}$ \\ 
                                               &     Overall & 33.11$^{+0.19}$ & 0.963$^{+0.02}$ & 31.41$^{+0.19}$ & 0.945$^{+0.03}$ \\ 
\midrule
NAFNet~\cite{chen2022simple}                   &           - & 33.71 & 0.966 & 31.22 & 0.943 \\ 
\multirow{2}{*}{NAFNet-MH-C~(\textbf{Ours})}   & Single head & 33.75$^{+0.04}$ & 0.967$^{+0.01}$ & 31.28$^{+0.06}$ & 0.944$^{+0.01}$ \\ 
                                               &     Overall & 33.82$^{+0.11}$ & 0.967$^{+0.01}$ & 31.33$^{+0.11}$ & 0.944$^{+0.01}$ \\ 
\bottomrule
\end{tabular}%
\end{table*}

\textbf{How does head combination influence the results?} The results in section 4.3 show that head combination improves the model performance while halving the parameters and computations of the output layer. Figure~\ref{fig:psnr_heatmap} plots the PSNR scores of all pairs of heads for the 4-head and 5-head model respectively. We can observe that the single head output on the diagonal performs worse, while the pair with different heads performs better. A possible explanation is that combining two different heads aggregates more information than a single head. The maximization operation allows for positive feedback during model training, which in turn results in better performance for the combination of different heads and worse performance for the single head. We further observe that There is no intersection between the best-performed pair of heads, i.e. pair $h0\&h1$ and $h2\&h3$. For $h4$ in the 5-head model, there is no other head to form a pair with it, so it performs moderately well with all other heads.

\textbf{Why do multi-head models achieve better $P\!S\!N\!R_{single\mbox{-}head}$ than their single-head counterparts?} We hypothesize that in deblur datasets like GoPro, the majority of labels are exactly the middle frame, and are deterministic when blurry input is given. But other parts of labels shift relative to the actual middle frames. When the single-output network is trained to fit such pairs, large losses occur, which may have a negative impact on training. However, the shifted label is still one feasible solution. Therefore, the proposed multi-outputs model allows a portion of the outputs in charge of restoring such pairs, making the model benefit from these samples while avoiding the negative effects of shifts on the output layer. So the multi-head models have a more stable training process and achieve better $P\!S\!N\!R_{single\mbox{-}head}$.

\subsection{4.5\quad Motion Deblurring Results}

In this section, we train our multi-outputs model on GoPro and evaluate it on GoPro and HIDE datasets. Table~\ref{tab:result} shows the results compared to the baseline single-output model and other approaches. The multi-head model with the head combination is represented with suffix \textit{-MH-C}. 

In details, the PSNR on GoPro of MPRNet, HINet, Restormer and NAFNet are improved by -0.01dB, 0.06dB, 0.08dB, 0.04dB, respectively. The PSNR on HIDE of MPRNet, HINet, Restormer and NAFNet are improved by -0.01dB, 0.08dB, 0.14dB, 0.06dB, respectively. The three-stage output scheme of MPRNet limits gains of its $P\!S\!N\!R_{single\mbox{-}head}$. The metric $P\!S\!N\!R_{overall}$ achieves 0.1dB gains on average, which means that our multi-outputs model is able to generate sharper images. When one of the generated sharp images matches the label, the PSNR gets higher.

\section{5\quad Conclusion}

Image deblurring is an ill-posed problem, and there are many feasible solutions for a blurry image. In this paper, we propose a multi-head outputs method to learn the distribution of these solutions. The method can be easily adapted to existing image restoration models. With the proposed multi-head outputs extension, as well as the corresponding loss function, the model is able to learn the distribution of sharp images given blurry input, instead of its expectation. In particular, the feasible solutions are aggregated into multiple clusters, and the objective of the training is to minimize the within-cluster sum of squared error. For implementation, the model outputs multiple sharp images, and only the head with the smallest loss is used for back-propagation. To further improve the utilization of each head, we propose a head combination method. The outputs are combined in pairs to obtain the extended outputs. The experiments on multiple models and datasets show that not only does the best overall PSNR metric exceed the baselines, but also the PSNR of each head is comparable to or even better than them. By extending the NAFNet, the proposed method surpasses the SOTA on the GoPro dataset.


\bibliography{aaai23}

\end{document}